# Self-paced Data Augmentation for Training Neural Networks

Tomoumi Takase[1*]   takase-tomoumi@aist.go.jp

Ryo Karakida[1]   karakida.ryo@aist.go.jp

Hideki Asoh[1]   h.asoh@aist.go.jp

[1] Artificial Intelligence Research Center, National Institute of Advanced Industrial Science and Technology, Tokyo, Japan

*Corresponding author

**ABSTRACT**

Data augmentation is widely used for machine learning; however, an effective method to apply data augmentation has not been established even though it includes several factors that should be tuned carefully. One such factor is sample suitability, which involves selecting samples that are suitable for data augmentation. A typical method that applies data augmentation to all training samples disregards sample suitability, which may reduce classifier performance. To address this problem, we propose the self-paced augmentation (SPA) to automatically and dynamically select suitable samples for data augmentation when training a neural network. The proposed method mitigates the deterioration of generalization performance caused by ineffective data augmentation. We discuss two reasons the proposed SPA works relative to curriculum learning and desirable changes to loss function instability. Experimental results demonstrate that the proposed SPA can improve the generalization performance, particularly when the number of training samples is small. In addition, the proposed SPA outperforms the state-of-the-art RandAugment method.

## 1. Introduction

Various techniques are used to improve the learning performance of neural networks. For example, data augmentation, which increases the number of samples by adding deformation to the original samples, is used in various applications, e.g., image classification and speech recognition [1, 2].

However, a significant limitation of data augmentation is that an appropriate method to apply this technique has not yet been established, even though it involves several factors that require careful tuning. Sample suitability is one such factors. In practical application, data augmentation is typically applied to all training samples regardless of sample suitability; however, this can reduce classifier performance. Thus, in this study, we focused on this problem because, to the best of our knowledge, it has not been addressed in previous studies.

In this paper, we propose the simple and practical self-paced augmentation (SPA) method to automatically and dynamically select suitable samples for data augmentation while training a neural network. The proposed method is easy to use and effective model training. We experimentally confirm the effectiveness of the proposed SPA in experiments using several typical benchmark datasets, including MNIST [3], CIFAR-10 [4], and Tiny ImageNet [5]. In addition, we compare the proposed SPA to the state-of-the-art RandAugment method [6]. We provide a detailed investigation of the effect of SPA on training performance and a simple comparison of prediction accuracy.

The remainder of this paper is organized as follows. In Section 2, we introduce related work, and, in Section 3, we describe our motivations and the procedure of the proposed SPA method. In Section 4, we report experimental results obtained for several numbers of samples, augmentation techniques, and datasets. In Section 5, we discuss two advantages of SPA relative to curriculum learning [7] and desirable changes in loss function instability. Finally, we conclude the paper in Section 6, including a summary of potential future work.

## 2. Related Work

In addition to typical data augmentation operations with affine transformation, e.g., flipping and rotation, several effective techniques have been proposed. For example, mixup [8] generates a new sample from linear interpolation of two samples using a mix rate sampled from a beta distribution. In addition, the cutout [9] and random erasing [10] methods apply masks to parts of images, and the random image cropping and patching (RICAP) [11] method randomly crops four images and then patches to generate a new image.

Recently, reducing the heuristics of data augmentation has attracted increasing attention. For example, the AutoAugment method [12] searches appropriate data augmentation policies using reinforcement learning. Although this provides high generalization performance, parameter search using reinforcement learning prior to training requires significant time. Lim et al. [13] reported that the parameter search requires 15,000 GPU-hours on the ImageNet [5] dataset. They proposed the Fast AutoAugment method, which reduced the time cost to 450 GPU-hours. However, the computation time remains considerably longer than that of conventional augmentation (CA) methods; thus, the Fast AutoAugment is not considered an easy replacement for CA methods.

Dynamically optimizing hyperparameters during training is effective relative to reducing computation time and constructing a practical method. Previous studies have addressed dynamic optimization of hyperparameters for neural networks other than data augmentation. For example, a previous study proposed a method to determine the learning rate dynamically such that the loss function on training data is minimized [14]. Other studies have proposed methods to dynamically tune the optimizer when training using meta-learning, which is a method to learn how to learn base training from higher-order training [15, 16]. Note that data augmentation should also be tuned dynamically during training based on some robust strategy.

RandAugment [6] randomly and dynamically selects augmentation techniques during training using a smaller search space than AutoAugment. As a result, this method accelerates training and significantly reduces computation compared to AutoAugment. However, this method only selects the suitable data augmentation techniques and tunes the hyperparameters in those methods, i.e., this method does not address the sample suitability problem. Identifying suitable samples for data augmentation and applying augmentation to those samples is significant relative to improving generalization performance. Therefore, we addressed the sample suitability problem in this study.

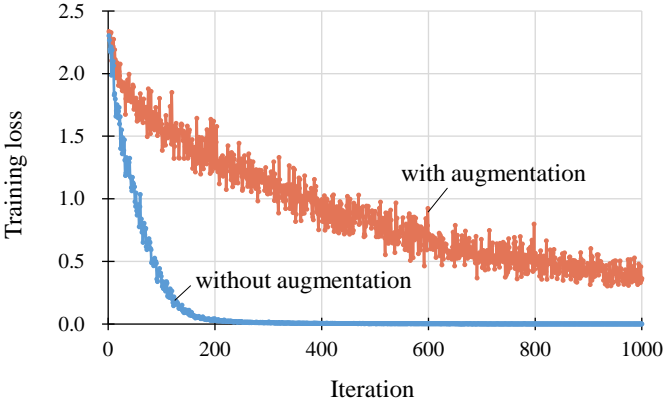

Fig. 1 Training loss transitions when data augmentation was and was not used. Here, a small CNN was trained on the CIFAR-10 dataset. Flip and random erasing data augmentation techniques were applied to all training samples. This result is part of the results shown in Fig. 6.

## 3. Proposed Self-paced Augmentation

### 3.1 Motivation

As described in the Section 1, data augmentation is conventionally applied to all training samples consistently during the entire process training. This is natural; however, this does not necessarily improve training performance because it does not consider sample suitability. This can be intuitively understood in Fig. 1, where the training loss transition when data augmentation was used is greater than that when data augmentation was not used. Thus, effective samples for data augmentation must be identified according to changing training conditions.

To determine sample suitability, we focused on the following two points. Here, one perspective we considered is curriculum learning [7], which is a strategy that transitions training from easy to difficult samples in a gradual manner. This is discussed and verified in detail in Sections 5.1 and 5.2. We also considered loss function behavior. As shown in Fig. 1, training loss was high and unstable when data augmentation was applied, i.e., the training loss fluctuated significantly in each iteration. This training loss trend does not work effectively relative to realizing high generalization performance, which is discussed and verified in detail in Sections 5.3 and 5.4. We designed the proposed SPA such that training is performed well in light of these considerations.

### 3.2 Procedure

Differing from conventional application of data augmentation, which is applied to all training samples, the proposed SPA method dynamically determines and selects samples that are suitable for data augmentation

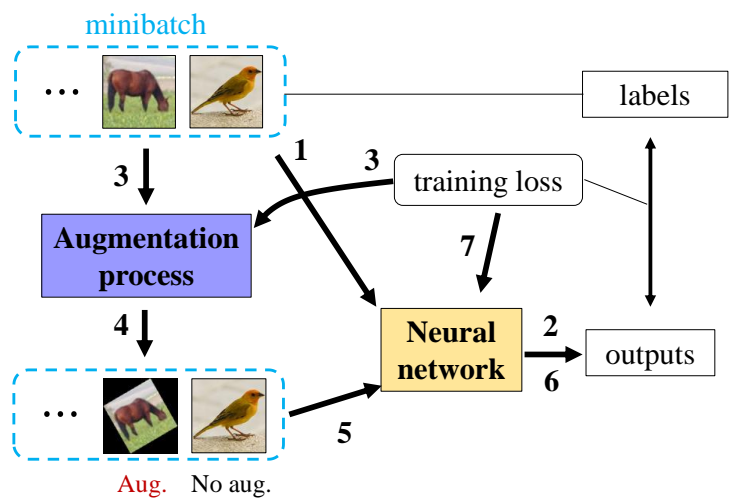

Fig. 2 SPA training procedure. The numbers beside the arrows indicate the order of the process.

```
Algorithm 1 SPA Training
1: Input:
       (x, y)
2: Output:
       θ
3: l ← Loss(f(x; θ), y)
4: for n from 0 to N − 1 do
5:     if l_n ≥ λ then
6:         (x_n, y_n) ← DA(x_n, y_n)
7:     end if
8: end for
9: l ← Loss(f(x; θ), y)
10: θ ← g(x; θ; l)
11: return θ
```

during training. The procedure of the proposed SPA is shown in Fig. 2.

The samples are input to a neural network prior to applying data augmentation to samples. Then, training losses are calculated using the outputs and labels, and are used for the augmentation process. Note that the weights of the neural network are not updated at this time. Here, data augmentation is applied to only samples that produce training loss greater than predetermined threshold $\lambda$. $\lambda$ is fixed in this method; however, the set of samples selected for data augmentation dynamically changes as training proceeds because the training loss for each sample changes during training. A detailed analysis of this process is given in Section 5. Then, regardless of whether data augmentation is applied, all samples in a minibatch are input to the neural network and the weights are updated. This overall process is performed for each minibatch.

Although Fig. 2 represents an example in which only rotation was applied to the samples, several data augmentation techniques, e.g., flip, rotation, and mixup [8], can also be applied to samples simultaneously. In addition, samples do not always change when data augmentation is applied because the degree of augmentation is determined randomly each time, e.g., if horizontal flip is applied, a sample is flipped with a probability of 0.5 and keeps unchanged with a probability of 0.5.

The procedure of this algorithm of the proposed method is outlined in Algorithm 1, where $f$ and $g$ are forward and back propagation, respectively, Loss is the loss function, $DA(x_n, y_n)$ is the sample to which data augmentation was applied, and $\theta$ is the neural network parameters. Although the same samples are input twice to the neural network in a single minibatch, only forward propagation is performed in the first input, which does not require significant computation time.

## 4. Experiments

### 4.1 Experimental Setup

In our experiments to evaluate the proposed SPA method, we focused on investigating the sample suitability problem in data augmentation, which is a new problem not considered by existing methods. Therefore, the main target of comparison was the general use of data augmentation, which augments all training samples. Thus, we compared the performances of the proposed SPA method to that of CA, which augments all training samples, and training with no augmentation (NA), which correspond to $\lambda = 0$ and $\lambda = \infty$ in the proposed SPA, respectively. For SPA, we considered several values for threshold parameter $\lambda$ (e.g., 0.01, 0.1, and 1).

Acquiring a large number of training samples is often difficult in terms of cost and availability. Data augmentation is used to train a model with a sufficient number of samples in such cases; thus, we focused on verifying data augmentation performance on a small number of training samples. In addition, the sample suitability problem addressed by the proposed SPA method may not be overly problematic in experimental setups with a large number of training samples. However, this is certainly relevant in setups with a small number of samples. Here, in addition to the original datasets with all full training samples, we used datasets prepared by randomly extracting part of the training samples from the original datasets.

To demonstrate the effectiveness of the proposed SPA on various datasets, we used several data augmentation techniques on several image benchmarks for machine learning, i.e., the CIFAR-10 and CIFAR-100 [4], MNIST [3], Fashion-MNIST [17], SVHN [18], STL-10 [19], and Tiny ImageNet [5] datasets. Note that training was always performed with a batch size of 100. In addition, these datasets were used in previous data augmentation study [12], and we followed that experimental setup.

Here, the initial weights were sampled from uniform distribution, and we used the Adam optimizer [20] with an initial learning rate of 0.001. The small CNN was trained for 1,000 epochs, and other CNNs were trained for only 200 epochs. The test accuracy on the original test data was calculated for each epoch. The best test accuracies obtained during training were compared, and we evaluated the mean and standard error for five trials with different initial weight values when a small number of training samples was used. Note that data augmentation was not applied to the test data.

We also conducted experiments on several data augmentation techniques. With the flip method, horizontal and vertical flips were applied independently at a probability of 0.5. For cropping, 80% of each image was randomly cropped and enlarged to the original size. With translation, each image was translated with an offset of $tx/128$, where $t$ is a value in the range $[-25, 25]$ and the image size is $x \times x$. With rotation, images were rotated using a random value in the range $[0, 360]$. Then, the images were downsized appropriately. In cutout [9], the mask size was set to 0.5, and, in random erasing [10], the mask size was set to a random value from 0.02 to 0.4 of the image size, and the aspect ratio was selected randomly in the range $[0.3, 3]$. In mixup [8] and RICAP [11], parameter $\alpha$ of the beta distribution to determine the mix rate was set to 1.

We compared the generalization performance of SPA to that of the conventional methods in four experiments (Sections 4.2–4.5). In the first experiment, we investigated the effects of the number of training samples. Here, we used a convolutional neural network (CNN) with a [conv-conv-pool-conv-conv-pool-fc] structure, where conv is a $3 \times 3$ convolution layer, pool is a $2 \times 2$ pooling layer, and fc is a fully connected layer (referred to as "small CNN" in the experimental results). In this CNN, batch normalization [21] and the rectified linear unit [22] were inserted after each layer, with the exception of the final layer, where the softmax function was used. We trained the model with flip or translation techniques on the CIFAR-10 or MNIST datasets.

In the second experiment, we used a small number of training samples. Here, we evaluated several data augmentation techniques on the CIFAR-10, Fashion-MNIST, and SVHN datasets and only used the flip and RICAP methods (which demonstrated high accuracy on the CIFAR-10 dataset) on the STL-10, CIFAR-100, and Tiny-ImageNet datasets. Considering the number of classes, we used 10,000 samples for CIFAR-100, 20,000 samples for Tiny-ImageNet, and 1,000 samples for the other datasets. For the models, we used ResNet34 [23] for CIFAR-100 and Tiny ImageNet, and we used the small CNN for the other models.

In the third experiment, we trained WideResNet28-10 [24] using various data augmentation techniques on the full CIFAR-10, Fashion-MNIST, and SVHN datasets.

In the fourth experiment, we compared the proposed SPA method to RandAugment [6], which is a dynamic augmentation method. Although it has demonstrated excellent performance on the full datasets and large CNNs with well-tuned hyperparameters, the effect on a small number of datasets has not been investigated sufficiently. Thus, we compared the performance of SPA to that of RandAugment on 1,000 training samples and full datasets. Here, the CIFAR-10, Fashion-MNIST, and SVHN datasets were used in to facilitate the comparison of these methods. With RandAugment, we selected and applied one data augmentation technique dynamically from 14 basic techniques, e.g., equalizing the image histogram and inverting all pixels above a threshold value of magnitude according to the settings used in the original study [6], and we used several values $(0, \cdots, 10)$ for the magnitude of data augmentation.

Table 1: Best test accuracy (%) when training the small CNN using several number of training samples. Values in parentheses are the standard error. Numbers in bold indicate the highest accuracy, and those in underline indicate considerably greater values than other methods. A data augmentation is regarded as being unsuitable when the accuracy for CA is considerably smaller than that for NA. "Number" denotes the number of training samples, and "trans." denotes the translation method.

| Dataset | Number | Aug. | SPA ($\lambda = 0.1$) | CA ($\lambda = 0$) | NA ($\lambda = \infty$) | Unsuitable |
|---|---|---|---|---|---|---|
| CIFAR-10 | 100 | flip | **30.87 ($\pm$0.33)** | 30.84 ($\pm$0.25) | 28.52 ($\pm$0.56) | |
| | | translation | 32.28 ($\pm$0.48) | **33.00 ($\pm$0.48)** | 28.52 ($\pm$0.56) | |
| | 500 | flip | **45.09 ($\pm$0.21)** | 43.84 ($\pm$0.30) | 42.23 ($\pm$0.36) | |
| | | translation | 48.29 ($\pm$0.59) | **49.11 ($\pm$0.75)** | 42.23 ($\pm$0.36) | |
| | 1000 | flip | **52.78 ($\pm$0.38)** | 50.19 ($\pm$0.35) | 48.47 ($\pm$0.49) | |
| | | translation | **56.96 ($\pm$0.42)** | 56.16 ($\pm$0.28) | 48.47 ($\pm$0.49) | |
| | 5000 | flip | **66.44 ($\pm$0.41)** | 63.51 ($\pm$0.39) | 64.01 ($\pm$0.19) | |
| | | translation | **72.80 ($\pm$0.18)** | 70.35 ($\pm$0.13) | 64.01 ($\pm$0.19) | |
| | 10000 | flip | **69.86 ($\pm$0.14)** | 68.10 ($\pm$0.09) | 69.28 ($\pm$0.35) | ✓ |
| | | translation | **75.98 ($\pm$0.11)** | 75.08 ($\pm$0.15) | 69.28 ($\pm$0.35) | |
| | (all) 50000 | flip | 77.71 ($\pm$0.13) | 77.70 ($\pm$0.12) | **78.32 ($\pm$0.11)** | ✓ |
| | | translation | 83.02 ($\pm$0.07) | **83.11 ($\pm$0.18)** | 78.32 ($\pm$0.11) | |
| MNIST | 100 | flip | **85.71 ($\pm$0.65)** | 74.25 ($\pm$0.90) | 84.12 ($\pm$0.92) | ✓ |
| | | translation | **92.14 ($\pm$0.18)** | 90.02 ($\pm$0.42) | 84.12 ($\pm$0.92) | |
| | 500 | flip | **96.81 ($\pm$0.11)** | 89.82 ($\pm$0.14) | 96.69 ($\pm$0.02) | ✓ |
| | | translation | **97.72 ($\pm$0.05)** | 97.20 ($\pm$0.07) | 96.69 ($\pm$0.02) | |
| | 1000 | flip | 97.77 ($\pm$0.04) | 93.36 ($\pm$0.02) | **97.77 ($\pm$0.10)** | ✓ |
| | | translation | **98.52 ($\pm$0.02)** | 98.23 ($\pm$0.04) | 97.77 ($\pm$0.10) | |
| | 5000 | flip | **98.88 ($\pm$0.05)** | 96.91 ($\pm$0.05) | 98.80 ($\pm$0.04) | ✓ |
| | | translation | **99.15 ($\pm$0.02)** | 99.00 ($\pm$0.01) | 98.80 ($\pm$0.04) | |
| | 10000 | flip | **99.27 ($\pm$0.01)** | 97.86 ($\pm$0.03) | 99.18 ($\pm$0.04) | ✓ |
| | | translation | **99.40 ($\pm$0.02)** | 99.21 ($\pm$0.01) | 99.18 ($\pm$0.04) | |
| | (all) 60000 | flip | **99.58 ($\pm$0.02)** | 98.76 ($\pm$0.06) | **99.58 ($\pm$0.04)** | ✓ |
| | | translation | 99.60 ($\pm$0.01) | **99.62 ($\pm$0.01)** | 99.58 ($\pm$0.04) | |

## 4.2 Effects of Number of Samples

The results shown in Table 1 indicate that the proposed SPA method tends to be effective when the number of samples is relatively small. In addition, the proposed method outperformed CA. In addition, the accuracy values obtained by the CA method were always worse than those of NA in the experiments with the MNIST dataset and the flip method because numbers can be confused with other numbers when flip is applied; thus, flip is unsuitable for images of such digits. This demonstrates that data augmentation does not always improve generalization performance because some augmentation techniques are not suitable for particular datasets. In addition, knowing the suitability of data augmentation for each dataset in advance is often difficult, as shown in the subsequent experiments. However, even in such cases, the accuracies of the proposed SPA were close to or greater than those of NA. This implies that SPA can mitigate deterioration in generalization performance when unsuitable augmentation is applied because the proposed SPA method does not always apply augmentation to all samples during training.

Table 2: Best test accuracy (%) when training the small CNN using various data augmentation techniques with a small number of samples from various datasets. We evaluated SPA using several $\lambda$ values, CA, and NA. Values in parentheses after accuracies are the standard error. 1k and 10k denote the number of training samples. Numbers in bold indicate the highest accuracy, and underlined values indicate considerably greater values than those of CA and NA.

| Dataset | Augmentation | SPA ($\lambda = 0.01$) | SPA ($\lambda = 0.1$) | SPA ($\lambda = 1$) | CA ($\lambda = 0$) |
|---|---|---|---|---|---|
| | flip | 51.62 (±0.36) | **52.78 (±0.38)** | 52.28 (±0.20) | 50.19 (±0.35) |
| | crop | 53.40 (±0.55) | **53.52 (±0.60)** | 51.66 (±0.51) | 52.71 (±0.47) |
| | translation | <u>57.63 (±0.29)</u> | 56.87 (±0.39) | 54.05 (±0.26) | 56.04 (±0.36) |
| | rotation | 40.67 (±0.37) | 47.72 (±0.48) | **<u>50.37 (±0.29)</u>** | 39.33 (±0.19) |
| CIFAR-10 (1k) | mixup [3] | 46.53 (±0.23) | **<u>50.47 (±0.22)</u>** | 49.94 (±0.18) | 46.45 (±0.19) |
| (NA: 48.47 (±0.49)) | cutout [4] | 49.15 (±0.21) | **<u>49.55 (±0.30)</u>** | 49.40 (±0.20) | 48.17 (±0.28) |
| | random erasing [5] | **<u>50.81 (±0.50)</u>** | 50.57 (±0.38) | 49.63 (±0.58) | 49.32 (±0.39) |
| | RICAP [6] | 53.80 (±0.28) | **<u>57.73 (±0.10)</u>** | 54.81 (±0.41) | 54.00 (±0.27) |
| | flip and crop | 55.31 (±0.44) | **<u>55.60 (±0.58)</u>** | 51.89 (±0.45) | 54.31 (±0.52) |
| | translation and rotation | 38.04 (±0.39) | 42.20 (±0.57) | **43.50 (±0.43)** | 36.36 (±0.58) |
| | flip | 84.82 (±0.19) | **84.83 (±0.16)** | 84.32 (±0.19) | 84.82 (±0.18) |
| | crop | 84.12 (±0.11) | **<u>84.37 (±0.14)</u>** | 83.64 (±0.23) | 80.00 (±0.21) |
| | translation | 83.65 (±0.16) | **<u>84.31 (±0.15)</u>** | 84.10 (±0.16) | 83.47 (±0.21) |
| | rotation | 79.46 (±0.26) | 82.61 (±0.21) | **<u>83.75 (±0.15)</u>** | 76.16 (±0.27) |
| Fashion-MNIST (1k) | mixup [3] | **<u>85.30 (±0.15)</u>** | 84.71 (±0.18) | 84.37 (±0.19) | 84.55 (±0.16) |
| (NA: 83.79 (±0.17)) | cutout [4] | 84.70 (±0.10) | 84.52 (±0.10) | 83.98 (±0.11) | **84.77 (±0.09)** |
| | random erasing [5] | **83.92 (±0.19)** | 83.78 (±0.18) | 83.83 (±0.17) | 83.81 (±0.20) |
| | RICAP [6] | 83.47 (±0.05) | **<u>85.72 (±0.19)</u>** | 85.04 (±0.17) | 83.47 (±0.17) |
| | flip and crop | 84.12 (±0.16) | **<u>84.63 (±0.11)</u>** | 83.83 (±0.12) | 79.82 (±0.21) |
| | translation and rotation | 75.81 (±0.12) | 80.45 (±0.19) | **<u>82.49 (±0.19)</u>** | 73.36 (±0.23) |
| | flip | 75.58 (±0.41) | 77.20 (±0.33) | **<u>78.64 (±0.23)</u>** | 71.29 (±0.50) |
| | crop | 78.47 (±0.36) | **<u>80.13 (±0.23)</u>** | 79.53 (±0.21) | 73.90 (±0.26) |
| | translation | 78.01 (±0.53) | 80.09 (±0.36) | **<u>80.17 (±0.35)</u>** | 73.28 (±0.53) |
| | rotation | 69.26 (±0.48) | 76.14 (±0.25) | **<u>78.93 (±0.23)</u>** | 51.47 (±0.65) |
| SVHN (1k) | mixup [3] | 75.31 (±0.10) | 79.82 (±0.18) | **<u>79.89 (±0.24)</u>** | 75.87 (±0.34) |
| (NA: 78.76 (±0.38)) | cutout [4] | **<u>79.22 (±0.30)</u>** | 79.16 (±0.33) | 79.04 (±0.29) | 77.28 (±0.34) |
| | random erasing [5] | 78.75 (±0.39) | 78.69 (±0.39) | 78.75 (±0.37) | 78.65 (±0.37) |
| | RICAP [6] | 72.10 (±0.32) | 78.38 (±0.44) | **<u>80.39 (±0.31)</u>** | 72.32 (±0.65) |
| | flip and crop | 73.67 (±0.61) | 78.10 (±0.47) | **<u>79.12 (±0.38)</u>** | 66.18 (±0.84) |
| | translation and rotation | 50.15 (±0.47) | 71.66 (±0.72) | **<u>75.67 (±0.54)</u>** | 41.02 (±1.05) |
| STL-10 (1k) | flip | **<u>54.71 (±0.72)</u>** | 54.24 (±0.65) | 52.38 (±0.36) | 53.43 (±0.43) |
| (NA: 49.58 (±0.42)) | RICAP | 59.62 (±0.35) | **<u>61.04 (±0.38)</u>** | 52.88 (±0.48) | 59.99 (±0.38) |
| CIFAR-100 (10k) | flip | 38.19 (±0.25) | **<u>40.08 (±0.21)</u>** | 38.86 (±0.11) | 38.64 (±0.29) |
| (NA: 32.62 (±0.27)) | RICAP | 50.03 (±0.76) | **<u>51.81 (±0.39)</u>** | 47.18 (±0.36) | 49.60 (±0.46) |
| Tiny-ImageNet (20k) | flip | 30.53 (±0.20) | 30.33 (±0.21) | **<u>31.58 (±0.44)</u>** | 30.60 (±0.21) |
| (NA: 25.10 (±0.57)) | RICAP | **<u>36.99 (±0.54)</u>** | 36.76 (±0.55) | 35.48 (±0.44) | 35.93 (±0.62) |

### 4.3 Experiments on Reduced Datasets

The results obtained with a small number of training samples on the various datasets are shown in Table 2. The results verify that the proposed SPA method outperformed CA and NA in most cases. Overall, high accuracy was generally achieved when $\lambda$ was set to 0.1.

Knowing the suitability of augmentation in advance is often difficult. For example, the rotation technique and its parameter value we used did not demonstrate high performance overall, and CA showed significantly reduced accuracy compared to NA. However, the proposed SPA tended to demonstrate high performances when the rotation technique was used. In addition, although cases that combined two augmentation techniques, e.g., flip and crop, did not demonstrate higher accuracy than when a single method was applied, e.g., only crop, the proposed SPA was effective in such cases.

### 4.4 Experiments on Full Datasets

The results obtained using various data augmentation techniques on the full CIFAR-10, Fashion-MNIST, and SVHN datasets are given in Table 3. As shown, the proposed SPA outperformed CA and NA in several cases; however, it did not always outperform these methods. When training with full datasets, a sufficient number of training samples were used; thus, the performance of SPA could be comparable with that of the CA and NA methods. Although the effectiveness of SPA is obviously smaller compared to cases with a small number of training samples, it is possible that training models using the proposed SPA will yield better performance than CA. In particular, as demonstrated in the results for the translation and rotation techniques, SPA still modified performance deterioration when unsuitable data augmentation techniques were applied.

Table 3: Best test accuracy (%) when training WideResNet28-10 using all samples from CIFAR-10, Fashion-MNIST, and SVHN datasets. We evaluated SPA using several $\lambda$ values, CA, and NA. Numbers in bold indicate the highest accuracy.

| Dataset | Augmentation technique | SPA ($\lambda = 0.01$) | SPA ($\lambda = 0.1$) | SPA ($\lambda = 1$) | CA ($\lambda = 0$) |
|---|---|---|---|---|---|
| CIFAR-10 (NA: 92.19) | mixup [3] | **91.76** | 91.07 | 90.71 | 91.67 |
| | cutout [4] | 91.66 | 90.59 | 90.53 | **91.98** |
| | random erasing [5] | 90.65 | **91.01** | 90.82 | 90.92 |
| | RICAP [6] | **93.98** | 92.92 | 90.88 | **93.98** |
| | flip and crop | **94.27** | 93.22 | 90.93 | 91.93 |
| | translation and rotation | 86.75 | **90.61** | 88.99 | 72.53 |
| Fashion-MNIST (NA: 93.99) | mixup [3] | 94.21 | 94.00 | **94.41** | 94.28 |
| | cutout [4] | 94.52 | 94.03 | 94.08 | **94.82** |
| | random erasing [5] | **94.21** | 94.02 | 94.11 | 94.05 |
| | RICAP [6] | 94.56 | **94.77** | 94.48 | 94.53 |
| | flip and crop | 94.38 | **94.83** | 94.67 | 94.61 |
| | translation and rotation | 93.75 | 93.93 | **94.27** | 92.97 |
| SVHN (NA: 96.48) | mixup [3] | **96.77** | 96.63 | 96.65 | 96.71 |
| | cutout [4] | 96.70 | 96.34 | 96.45 | **96.74** |
| | random erasing [5] | 96.50 | 96.48 | **96.52** | 96.42 |
| | RICAP [6] | 97.35 | 97.29 | 97.12 | **97.50** |
| | flip and crop | **97.34** | 96.78 | 97.27 | 94.04 |
| | translation and rotation | 95.49 | 96.29 | **96.84** | 87.23 |

**4.5 Comparison with RandAugment**

We then compared the test accuracies of the proposed SPA to those of the RandAugment method [6]. With 1,000 samples from the CIFAR-10 dataset, the test accuracies of RandAugment were always less than 40.00%, and the highest accuracy was 37.74% with an augmentation magnitude of 5. This accuracy is much less than that of the proposed SPA (and even CA or NA), as shown in Table 2. Thus, the proposed SPA method was more effective than RandAugment with a small number of training samples in our experimental setting.

In addition, we evaluated the test accuracies when training using all samples from several datasets. The test accuracies for RandAugment on the CIFAR-10, Fashion-MNIST, and SVHN datasets were 96.39%, 94.06%, and 96.34%, respectively. The best results obtained by the proposed SPA were 94.27%, 94.83%, and 97.34%, respectively (Table 3), and the RandAugment method obtained the best accuracy on the CIFAR-10 dataset; however, the proposed SPA obtain higher accuracy on the Fashion-MNIST and SVHN datasets. Thus, we consider that the proposed SPA is also effective on full datasets, even though it is the most effective with a small number of training samples. Against our expectations, the proposed SPA demonstrated high performance overall even though fixed data augmentation techniques were during training, whereas the RandAugment method randomly selected one of 14 techniques for iteration.

**5. Discussion**

Here, we discuss why the proposed SPA method performed well in consideration of two well-established training strategies, i.e., curriculum learning [7] and desirable changes to loss function instability [25]. First, we introduce each strategy and describe how SPA is based on these strategies. We then demonstrate the effect of these strategies on SPA through experiments.

**5.1 Curriculum Learning**

Curriculum learning is a training strategy based on the difficulty of samples [7], where training proceeds from easy to difficult samples in a gradual manner to approximate human learning. Baby-step learning [26], which is an improved curriculum learning technique, trains a model using previously used samples, whereas curriculum learning replaces easy samples with difficult samples during training.

These methods require that samples be grouped manually into several sets according to their difficulty, which is achieved using prior knowledge that depends on the data. In contrast, self-paced learning (SPL) [27] automatically creates a boundary surface between difficult and easy samples. Here, samples that produce large losses are considered difficult samples (and vice versa). Thus, sample difficulty is flexible in SPL, although it is fixed during training in the original curriculum learning.

When $v^{(i)}$ denotes $N$-dimensional values with 1 or 0 that represent whether $N$ samples are used for training at the $i$th epoch during training with SPL, $v^{(i+1)}$ can be expressed as follows:

$$v^{(i+1)} = \min_{v^{(i)}} \left( \sum_{n=0}^{N-1} v_n^{(i)} L(y_n, f(x_n, \theta)) - \eta \sum_{n=0}^{N-1} v_n^{(i)} \right), \quad (1)$$
$$\text{s.t. } v^{(i)} \in [0,1]^N,$$

where $x_n$ and $y_n$ are the inputs and label of the $n$th sample, respectively, $\theta$ represents the model parameters at that time, $f(x_n, \theta)$ is the predicted label, $L$ is the loss function, $N$ is the number of samples, $v_n^{(i)}$ is a value (1 or 0) that defines whether the $n$th sample was used for training at the $i$th epoch, and $\eta$ is a parameter to control the training rate. If each $v_n^{(i)}$ becomes 1 to minimize the second term, $\sum_{n=0}^{N-1} v_n^{(i)} L(y_n, f(x_n, \theta))$ in the first term increases. Thus, $v_n$ for samples with small $L(y_n, f(x_n, \theta))$, i.e., easier samples, tend to become 1 to minimize $E_A(\theta; v; \eta)$. Parameter $\eta$ is increased gradually during training to increase the number of samples used for training and to use more difficult samples.

Several studies have used or improved SPL in the context of neural networks [28, 29]. A common strategy in the above methods related to curriculum learning is training on easier samples in the early training stage, which is important to realize improved generalization and faster convergence [30–32]. In addition, Cirik et al. [33] demonstrated experimentally that curriculum learning on long short-term memory networks [34] improves significantly when smaller models are used and helps more when the amount of training data is limited.

### 5.2 Curriculum Learning in SPA

The proposed SPA method differs from the original curriculum learning because SPA uses difficult samples by applying data augmentation and easy samples in the early stage. In addition, SPA does not use difficult samples in the final stage. However, SPA applies data augmentation strategically based on a unique curriculum that partially uses the idea of original curriculum learning relative to the following two points;

thus, curriculum learning is one factor that constructs SPA rather than a comparative method.

The first point is to use easier samples for training in the early stage than CA uses. As shown in Fig. 1, applying data augmentation makes samples more difficult; thus, SPA can use easy samples for training by applying data augmentation to only difficult samples. Thus, the samples used for training in the early stage in SPA are easier than those in CA, which only uses difficult samples in the stage by applying data augmentation to all samples.

The other point is that sample selection for data augmentation in SPA is achieved using a procedure that is similar to that used in SPL. In SPA, sample difficulty is assessed using the loss function value when each sample is input to the neural network, and this technique is also employed in in SPL [27]. When $\boldsymbol{v}^{(i)}$ denotes $N$-dimensional values with 1 or 0 that represent whether data augmentation is applied to each of $N$ samples at the $i$th epoch during training with SPA, $\boldsymbol{v}^{(i+1)}$ can be expressed based on Eq. (1) as follows:

$$\boldsymbol{v}^{(i+1)} = \min_{\boldsymbol{v}^{(i)}} \left( \sum_{n=0}^{N-1} \left(1 - v_n^{(i)}\right) L\left(y_n, f(\boldsymbol{x}_n, \boldsymbol{\theta})\right) + \eta \sum_{n=0}^{N-1} v_n^{(i)} \right). \tag{2}$$

$$\text{s.t. } \boldsymbol{v} \in [0, 1]^N,$$

Here, $v_n^{(i)}$ tends to become 1 for a sample with large $L(y_n, f(\boldsymbol{x}_n, \boldsymbol{\theta}))$. Data augmentation is applied to samples with $v_n^{(i)} = 1$, and the number of such samples can be reduced by increasing $\eta$ during training.

Unlike Eq. (1), $\left(1 - v_n^{(i)}\right)$ weighs the first term, and the sign of the second term is changed. The difference in the first term comes from the fact that "easy" samples are used for training in SPL, whereas "difficult" samples are used for augmentation in SPA. The difference in the second term comes from the fact that the number of samples with $v_n^{(i)} = 1$ increases during training in SPL, whereas it decreases in

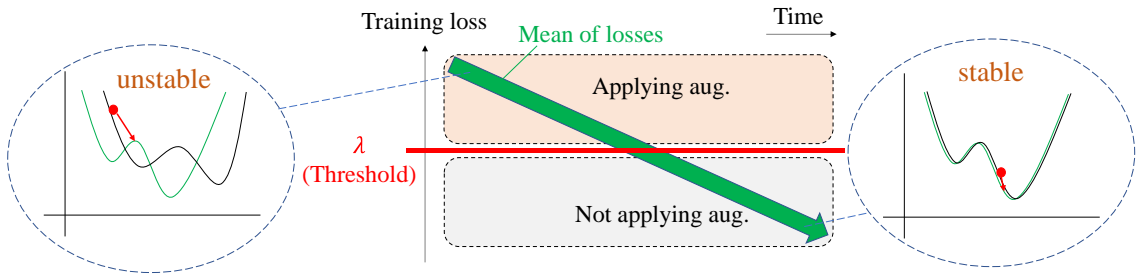

Fig. 3 Criterion by which data augmentation is applied in SPA. The concept of curriculum learning and desirable change of loss function instability is summarized. "Aug." denotes data augmentation.

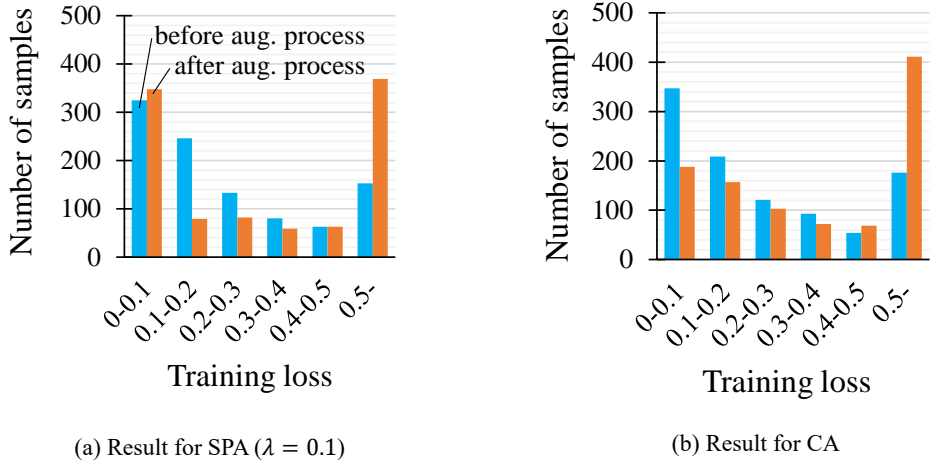

(a) Result for SPA ($\lambda = 0.1$)  (b) Result for CA

Fig. 4 Histograms of the number of training samples according to training losses. Numbers before and after the augmentation process at epoch 20 are shown for SPA and CA. Random erasing was used as the data augmentation technique.

SPA. While the proposed SPA method is considered to follow a well-established idea of SPL in terms of sample selection, the proposed method involves much more than simply using SPL, i.e., it is specifically designed for data augmentation.

In implementation, rather than using Eq. (2), we decided on $v^{(i+1)}$ by comparing $L(y_n, f(x_n, \theta))$ to $\lambda$, which is a hyperparameter. As shown in Fig. 3, a sample producing training loss that is greater than threshold $\lambda$ is considered a difficult sample, and data augmentation is applied to it (and vice versa). Here, appropriate $\lambda$ depends on the datasets and models; thus, it should be investigated by attempting several $\lambda$ values. As a whole, many difficult samples exist in the early stage; however, the number of difficult samples decreases as training proceeds because the training loss for each sample tends to decrease. Recall that the procedure of SPA in detailed in Algorithm 1.

We confirmed that the effect of curriculum learning appears in actual training. Here, we counted the numbers of samples according to training losses before and after applying data augmentation at epoch 20. The histograms of the numbers are shown in Fig. 4. When using SPA with $\lambda = 0.1$ (Fig. 4(a)), the number of samples with training loss less than 0.1 did not decrease by applying data augmentation because those samples were not augmented in SPA, whereas, with CA (Fig. 4(b)), the number of samples with training loss less than 0.1 decreased greatly. Thus, as expected, easy examples without data augmentation can be used in the early stage of SPA training.

In addition, we trained the small CNN with 1,000 samples from the CIFAR-10 dataset by applying data augmentation to randomly selected samples to demonstrate that applying data augmentation to difficult samples contributes to improved test accuracy. Here, the number of samples to which data augmentation was applied during training was the same as the number of samples with $v_n = 1$ in SPA. The best test accuracy was obtained when data augmentation was applied randomly with $\lambda = 0.1$ was 49.12%($\pm$0.35), whereas that for SPA with the same $\lambda$ value was 52.78%($\pm$0.38). This demonstrates that training in the early stage with easy samples without data augmentation is effective.

### 5.3 Loss Function Behavior

Although minimizing training loss is required to obtain high generalization performance, generally, the parameter with a small training loss does not always produce high generalization performance if it is in a poor minimum. Previous studies [35, 36] have indicated that generalization performance is degraded when the parameter is trapped in a sharp minimum.

In stochastic gradient descent, a minibatch is used, and the loss function with parameter $\boldsymbol{\theta}$ is expressed as $L = f(\boldsymbol{x}, \boldsymbol{\theta})$ using variable inputs $\boldsymbol{x}$. This can be considered as optimizing different loss functions in each time step; however, it can also be considered as optimizing a fixed average loss function with a parameter that moves stochastically [37]. Takase et al. [25] referred to this fluctuation in loss function from one step to another as loss function instability.

Intuitively, if we can control the instability and change the loss function from unstable to stable during training (Fig. 3), this will help to avoid becoming trapped in poor local minima, e.g., sharp minima, and facilitate stable convergence. In fact, Takase et al. [25] focused on the fact that a small minibatch size destabilizes the loss function, and they proposed a method that increases the batch size gradually during training. This facilitates searching a wide range of loss functions in the early stage, and then, the search range is reduced gradually, which results in improved generalization performance. This is similar to the idea of simulated annealing [38–40], although it controls the behavior of the parameter rather than the function.

### 5.4 Loss Function Behavior in SPA

The proposed SPA minimalizes the original loss function without data augmentation in the final training stage. However, it generalizes better than NA, which suggests that the transient dynamics of the proposed SPA in the early training stage will provide some advantage relative to finding better global minima.

With this expectation in mind, we confirm that small training loss and changing the loss function from instable to stable are observed in SPA-based training. First, we demonstrate that data augmentation increases training loss and destabilizes the loss function. Here, we trained the small CNN using several data augmentation techniques with the full CIFAR-10 dataset. The transition plotted in Fig. 5 shows that data augmentation increases training loss and destabilizes the loss function because the training loss increases eventually and changes more significantly for each minibatch compared to the no data augmentation case. Using both flip and random erasing provided the greater training loss and the most instable loss function.

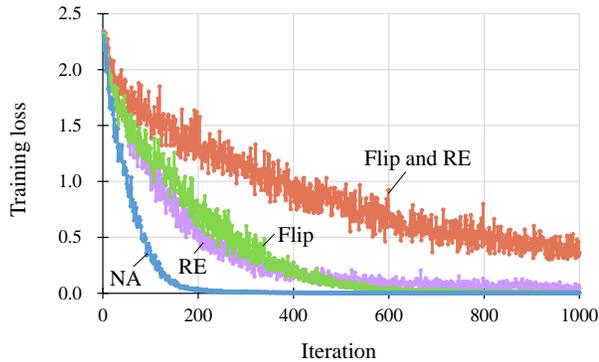

Fig. 5 Effects of data augmentation on training loss instability. RE denotes random erasing. In cases other than NA, data augmentation was applied to all training samples.

This result implies that we can control the value and instability of the loss function by devising how to apply data augmentation. Here, a desirable strategy is to apply data augmentation to many samples in the early stage and then gradually reduce the number of samples for data augmentation. This can greatly

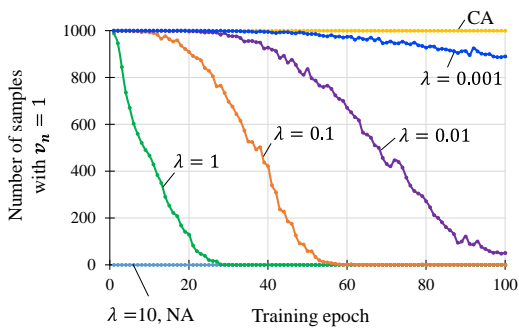

(a) Training for the small CNN on 1,000 samples

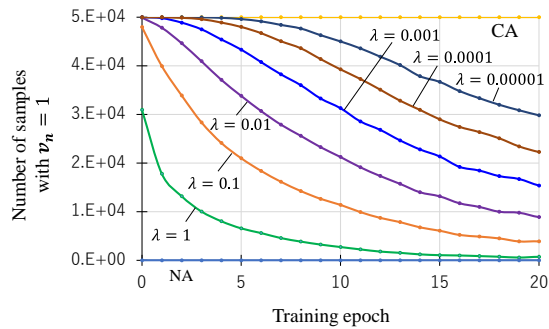

(b) Training for WideResNet28-10 on all samples

Fig. 6 Transitions of the number of samples with $v_n = 1$ during training. The flip augmentation technique was applied to the CIFAR-10 dataset.

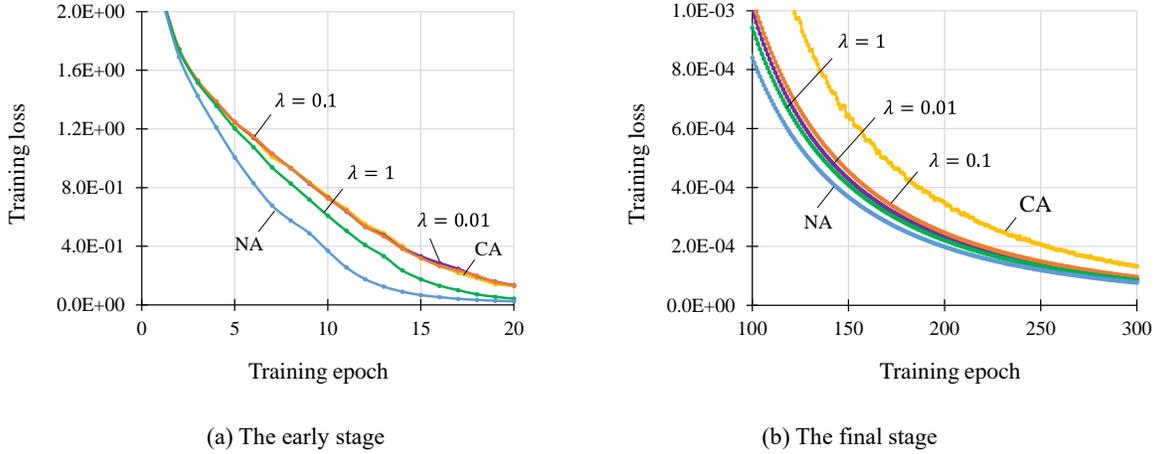

(a) The early stage  (b) The final stage

Fig. 7 Transitions of training loss for each minibatch. Effect of $\lambda$ on training loss was investigated. The small CNN was trained with 1,000 samples from the CIFAR-10 dataset.

reduce training loss and make the loss function unstable in the early stage and stable in the final stage. Equation (2), i.e., the equation for SPA, includes this desirable change in loss function instability and the effect of curriculum learning. As discussed in Section 5.2, the mean training losses tend to decrease during training because training is performed using gradient descent. Therefore, the number of samples for data augmentation tends to gradually decrease.

We confirmed this was achieved in actual training. Figure 6 plots the transitions of the number of samples with $v_n = 1$, i.e., samples to which data augmentation was applied, during the first one tenth in the trainings with the flip augmentation technique. These results demonstrate that the number of samples with $v_n = 1$ decreased gradually with appropriate $\lambda$ values, and smaller $\lambda$ values decreased the rate of reduction, approaching the transition for CA. Although samples with $v_n = 1$ were very few (or none) depending on the $\lambda$ value, this does not necessarily lead to reduced accuracy, as discussed in a previous study [41], which insisted that using less augmented data in the final training stage can refine the model and improve accuracy.

Figure 7 shows the sum of training losses for the minibatches when training with a small number of samples. Although the transition of SPA resembles that of CA in the early stage (Fig. 7(a)), it resembles that of NA in the final stage (Fig. 7(b)), and it became more similar to that of NA as the $\lambda$ values increased. In reference to the literature [25], we also evaluated training loss instability according to the variance in training loss. Figure 8 shows the transitions of the variance in training loss for each minibatch. Here, to

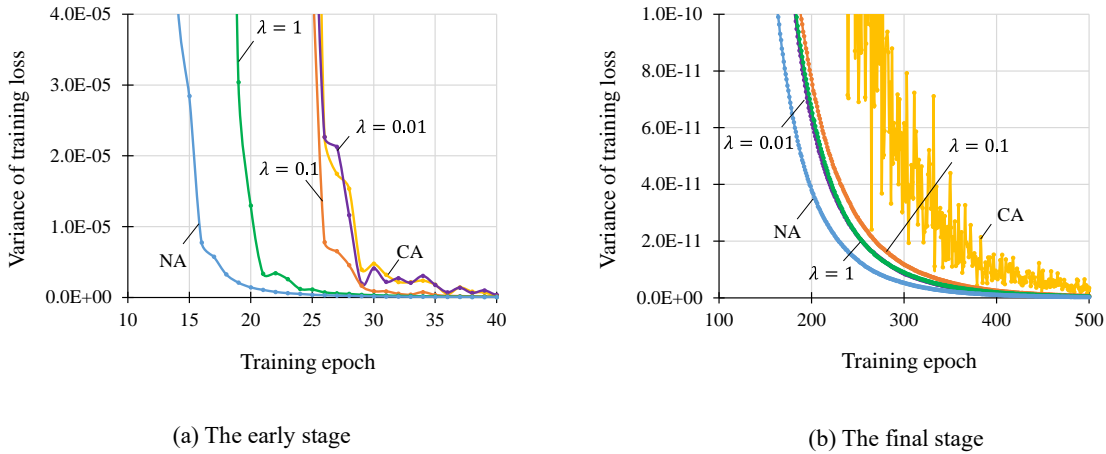

(a) The early stage

(b) The final stage

Fig. 8 Transitions of variance in training loss for each minibatch. The effect of $\lambda$ on the loss function instability is investigated. The small CNN was trained with 1,000 samples from the CIFAR-10 dataset.

facilitate reliable evaluation, training loss was measured at varying positions according to the neural network parameters rather than at a fixed position. Similar to the results shown in Figs. 6 and 7, SPA demonstrated transitions that are similar to those of CA and NA in the early and final stages, respectively. Thus, loss function instability changed from unstable to stable when training with SPA using a small number of training samples. The transient dynamics of SPA are expected to facilitate avoiding poor or sharp local minima (Section 5.3), although this may be just one of the factors of the proposed SPA method

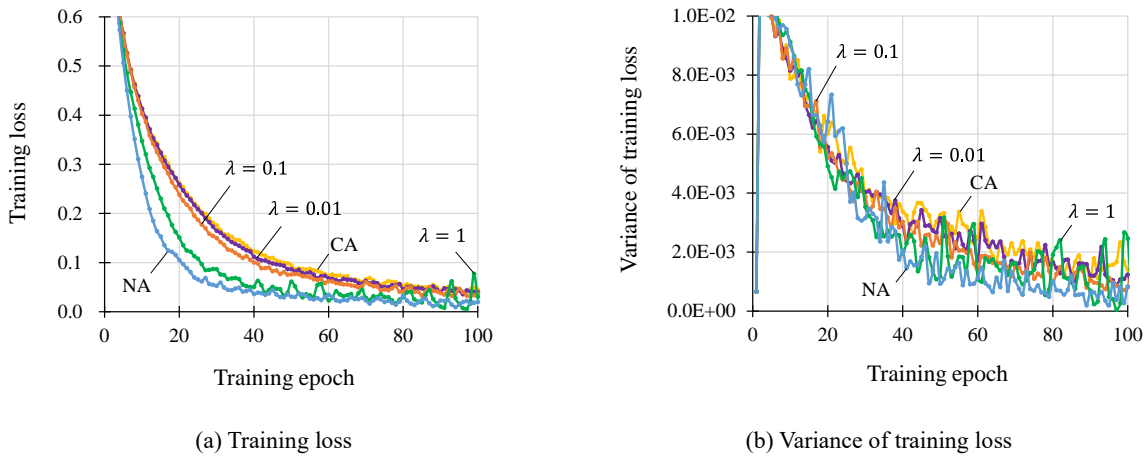

(a) Training loss

(b) Variance of training loss

Fig. 9 Training loss and transitions of the variance in training loss for each minibatch. The small CNN was trained with all samples from the CIFAR-10 dataset.

that enhances generalization performance.

We conducted the same experiments using all training samples. Figure 9 shows the transitions of the training loss and variance in loss for each minibatch when training the small CNN on the CIFAR-10 dataset. Trends that differ from those for a small number of training samples can be observed in these results. In the transition of training loss shown in Fig. 9(a), the loss for SPA is nearly the same as that of CA in the final stage. In the transition of the variance of training loss shown in Fig. 9(b), SPA demonstrates a similar trend with CA and NA. In addition, all methods showed large fluctuation. Therefore, the effect of SPA was small; thus, significant improvements to test accuracy cannot always be expected, as shown in Table 3. As SPA smoothly connects CA and NA, and exploits the advantages of both methods, it performs the most effectively in the conditions such as Figs. 7 and 8 that the trends in training loss of CA and NA are greatly different.

## 6. Conclusions

In this paper, we have proposed SPA, which dynamically selects samples for data augmentation. The proposed SPA method only applies data augmentation to samples that produce a training loss greater than fixed threshold $\lambda$, and this technique was designed based on curriculum learning and desirable changes in loss functions.

The results of extensive experiments on several datasets with different data augmentation techniques verify the effectiveness of the proposed SPA method, which outperformed both the CA and NA methods in most cases. In particular, SPA demonstrated better performance in two practical situations, i.e., when only a small amount of training data is obtained and when unsuitable data augmentation techniques are used. In addition, in such situations, the proposed SPA significantly outperformed the state-of-the-art RandAugment method. Our experimental results confirm that the effects of curriculum learning and desirable changes in loss functions are evident when training using the proposed SPA method.

Overall, we believe that we have addressed the problem of selecting suitable training samples for data augmentation, which enables more appropriate application of data augmentation techniques. However, several open issues remain relative to the heuristic settings of data augmentation. For example, the parameters of each data augmentation technique, e.g., mix rate in mixup, should be adjusted automatically and dynamically. In addition, although several $\lambda$ values were used to evaluate the proposed SPA in our experiments, and $v_n = 0.1$ tended to yield high accuracy, incorporating $\lambda$ optimization into a training system using SPA is expected to be more useful. Thus, in future, we plan to address these issues to improve the generalization performance of neural networks using data augmentation.